# A Faster Structured-Tag Word-Classification Method

**Min Zhang**
University of Melbourne
mzh@cs.mu.oz.au

**Abstract**

Several methods have been proposed for processing a corpus to induce a tagset for the sub-language represented by the corpus. This paper examines a structured-tag word classification method introduced by McMahon (1994). Two major variations (*non-random initial assignment of words to classes* and *moving multiple words in parallel*) together provide robust non-random results with a speed increase of 200% to 450%, at the cost of slightly lower quality than McMahon's method's average quality. Two further variations, *retaining information from less-frequent words* and *avoiding reclustering closed classes,* are proposed for further study.

## 1. Introduction

Many tasks in computational linguistics, whether they use statistical or symbolic methods, reduce the complexity of the problem by dealing with classes of words rather than individual words. The names of the classes are known as "tags". To determine a suitable set of tags (a "tagset") requires considerable linguistic expertise, and to create a lexicon by manually assigning tags to words is a very large task. Much research has been devoted to trying to find a fast reliable way of doing this automatically by processing large text corpora.

In this paper I focus on a method proposed by McMahon (1994) which produces a structured tagset of reasonable quality in reasonable time, and examine some variations to the basic algorithm. Two of these variations result in a considerable speed increase, with little reduction in quality.

The method is not tailored to English or any other particular language. It can be used to induce a tagset for any (sub-)language or domain-specific language which has no computational lexicon. Results are given for the application of the method to text generated from a small test grammar, a database of customer names, and an English novel.

## 2. Clustering words

In a domain where a sub-language is used, it is very unlikely that an existing tagset derived from general texts in the full language (e.g. the Penn Treebank tagset) will be useful. In sub-domains like those found in free-text fields in commercial databases (for example, names of customer), a quick glance at the corpora shows that the input strings seem to be mostly noun phrases rather than full sentences; there are no verbs; and nouns can be and should be sub-divided into multiple classes.

We know that some words share similar sorts of linguistic properties, thus they should belong to the same class. Some words have several functions, thus they could belong to more than one class. The questions are: What attributes distinguish one word from another? How should we group similar words together so that the partition of word spaces is most likely to reflect the linguistic properties of language? What meaningful label or name should be given to each word group? These questions constitute the problem of finding a tagset (or word classification) for a language or sub-language. At present, no method that exists can find the optimal word classification. However, researchers have been trying hard to find sub-optimal strategies which lead to useful classification.

Most automatic word-class clustering methods consider "hard" or "Boolean" word classes: a word is assigned wholly to <u>one (and only one)</u> class of a set of <u>distinct</u> or <u>discrete</u> classes. There is no overlap between one class and another. Of course, a word can be used in different syntactic categories. For example, words which behave only like nouns, only like verbs, or like either verbs and nouns would be expected to be assigned to three different "hard" classes by these methods. Pereira et al (1993) proposed a method for "soft" class clustering, grouping words "as *clusters c* with corresponding cluster membership probabilities *p(c/w)* for each word *w*". However, their paper does not go into inducing a tagset but only considers the special case of classifying nouns according to their distribution as direct objects of verbs.

## 2.1  Structured tags

One alternative to the creation of a set of distinct classes is to build a hierarchical structure of classes. Tags assigned to classes in such a structure are called "structured tags". The hierarchy can be represented as a tree. If we restrict this tree to being a binary tree, the structured tag is easily represented as a string of bits.

A clustering method that can be used to create structured tags is considered to be potentially more suitable because the structured tag offers more information about properties of word classes. With a non-structured tag, we are just told that a word class is different from another class, but we do not know <u>how great</u> the difference there is. With a hierarchical classification, we know that the closer the positions of the classes in the hierarchical structure, the closer the relationship between the classes. Two binary structured tags can be compared relatively efficiently by counting the number of leading identical bits. The greater this number, the more similar the two word classes are. That could be very useful in a parser that selects a rule by some heuristic that allocates a score based on matching word classes; the structured tag allows finer tuning by scoring <u>between</u> 0 and 1 rather than exactly 1 for a match and 0 for a non-match.

## 2.2  Previous work

There are two broad groups of word classification methods. One group works on a vector of feature values for each word; the clustering is then based on some measure of similarity between pairs of words. Finch & Chater (1992) use a feature vector of size 600 (number of occurrences of 150 frequent words in each of two positions on either side), with Spearman's Rank Correlation Coefficient as the similarity metric. Hughes & Atwell (1994a, 1994b) investigate eight clustering methods, and also evaluate several combinations of feature vectors and three methods of similarity measures. The combination of Ward's clustering method and the Manhattan distance metric was found to be the best.

In Schütze's work (1993, 1995), the feature vector of a word contains the numbers of shared neighbouring words at the left side and right side of this word, and the similarity between two words is taken as the cosine of their feature vectors. The problem of data sparseness is overcome by the application of a singular value decomposition (SVD) method.

Brill & Marcus (1992) use a feature vector containing the probabilities of neighbouring words (one dimension for one word). The distributional similarity score, which is based on probability distributions over environments, is calculated to compare two words. This is the sum of two "Kullback-Leibler distances" or "relative entropies" $D(P_1 \| P_2) + D(P_2 \| P_1)$.

The other group attempts to optimise some overall measure of "goodness of clustering" such as "average mutual information" or "perplexity". Brown et al (1992) partition a vocabulary of V words into C classes; the similarity between two adjacent classes is measured by mutual information I(*c1*, *c2*). They propose two algorithms; in their second algorithm, one for a vocabulary of a realistic size, the words in the vocabulary are arranged in order of frequency with the most frequent word first and each of the first C words is assigned to its own class. At the *k*th step of the algorithm, the (C+*k*)th most frequent word is assigned to a new class (C+1). The pair of classes is found for which the loss in average mutual information is least if they are merged. That pair of classes is merged, reducing the number of classes to C again.

Jardino & Adda (1993) proposed a simulated annealing method. The perplexity, which is the inverse of the probability over the entire text, is measured. The new value of the perplexity and a control parameter cp (Metropolis algorithm) will decide whether a new classification (obtained by moving only one word from its class to another, both word and class being randomly chosen) will replace a previous one. The computer run time for this method seems much faster than others; it took only seven hours on a PC to classify a vocabulary of 14,000 words into 120 classes. However they report the quality thus: "The classification often makes sense from a linguistic point of view". I would not regard "often" as sufficiently frequent.

Chang & Chen (1995) use what they describe as the same method as Jardino & Adda to classify Chinese words into classes. However it takes more than one week (201 hours) on a DEC 3000/500 AXP workstation to classify a vocabulary of 23,977 words into 200 classes, which is too costly to be feasible.

## 2.3  *Summary of previous work*

Running time seems to be a problem with all of the methods I have looked at. Only a few papers report timing information; and for these the running time seems very long, with the exception of course of Jardino & Adda. Time taken seems to be at least $O(n^3)$ where n is the number of classes being merged or otherwise worked on. If n is the number of words in the vocabulary, the time will be too great. A two-stage method like that proposed by Brill & Marcus (choosing a small number of the most frequently occurring words to process first and lower frequency words processed later by a different method) seems to be necessary.

As far as the quality of the various clustering methods is concerned, not much can be said. The hope is that methods like these can assist linguists to develop tagsets for (sub-domains of) languages. I believe that it is best left to the judgement of the linguist whether a tagset is adequate for the purpose in mind. Although Hughes & Atwell name this approach the "looks good to me" method, what they offer instead is only a method of comparing a clustering result to an existing tagged corpus, assuming one is already available. The result of the comparison depends heavily on the tagset used to tag the corpus. In fact the tagset they used is reduced (by them) for benchmark purposes to 19 tags, presumably using linguistic judgement to arrive at their "ideal of 19". Their method assumes that the existing tagset is adequate. However, if a novel clustering method discovers some difference in word distribution that was overlooked by the creators of the benchmark's tagset or tagger, the evaluation method cannot rank the result of the novel method better than the benchmark; closeness to the benchmark is "good" and divergence is "bad". Furthermore, it seems unfair to rate clustering methods on the basis of a test involving one language, one corpus, and a single heavily-reduced tagset. In my opinion, it would require an evaluation over multiple languages, corpora and tagsets to judge whether one clustering method is better than another.

None of the methods seem good enough for entirely automatic derivation of word classes. A procedure like that of Brill & Marcus where classes are reviewed by the analyst and words are assigned or re-assigned to classes would seem to be still needed.

## 3. McMahon's structured-tag clustering method

Work by McMahon (1994), summarised by McMahon & Smith (1995), uses average class mutual information (ACMI), like Brown et al., as the metric. However McMahon uses a totally different top-down word class splitting technique, which explicitly produces binary structured tags. The clustering is done this way: At the top level of the binary tree, it treats all words in a corpus as if they belonged to only two classes, then it attempts to find the "optimal" two-class word division. This determines for each word whether the first bit in its tag is 0 or 1. At the second level, the two classes are split into four, thus assigning the second bit in the tag, and so on. The split at each level is an iterative process. Each iteration finds the word which, if moved to the opposite class, would maximise the ACMI.

There can be no guarantee that the method will find the global maximum of ACMI. It is also important to note that the initial assignment of words to classes is random. This can mean that different runs of the algorithm can find different local maxima. The empirical evidence, shown in section 5.1, is that multiple local maxima can exist, especially when the data is scarce.

McMahon evaluated his results using Hughes's evaluation method. His system performed better than Finch & Chater's, but worse than Hughes's. After comparing the technique with those of other authors (Brown et al, Brill & Marcus, Schütze, Finch & Chater) and examining their strengths and weaknesses, McMahon suggests that a hybrid top-down and bottom-up system would be better than a single approach.

The computation time is said to be of order $O(sV^3)$ for tree height s and vocabulary size V. The step in the algorithm that is executed $sV^3$ times is (or should be) only a very simple step that moves a word from one class to another. However at each level the algorithm takes at least $O(V^2C)$ ACMI calculation steps, where C is the number of classes being considered at that level. Each step involves a logarithm calculation, which may be very expensive compared to the class-moving step. Even if not $O(V^3)$, the actual time taken may depend mostly on the ACMI calculation.

McMahon does not discuss the question of a suitable value for the maximum tree height s, or any other stopping condition for the algorithm. His implementation evidently uses a 16-bit tag, but it is possible that this is not prompted by a perceived need for 65,536 word-classes. McMahon & Smith (1995) say "we would like to construct a clustering algorithm which assigns a unique s-bit number to each word in our vocabulary", without saying why the number should be unique. For clustering applications which require human-comprehensible output, providing each word with a necessarily non-unique word-class out of a tagset of size (say) 256 to 1024 would seem adequate. An implementation is then free to use some other independent and simple numbering scheme for uniquely identifying words during the clustering. Attempting to maintain the tag as a unique number for each word would lead to considerable overhead in the inner loops of the algorithm.

## 4. Variations on McMahon's method

While implementing McMahon's structured-tag automatic classification method according to the descriptions of his algorithm, some possible variations became apparent; I have implemented these variations and compared the results with the basic method.

### 4.1 Non-random initial assignment of words to classes

At each level, each word must be assigned to one of two classes. In other words, the bit for that level for each word is either 0 or 1. McMahon's method makes the initial assignment at random. Thus, the probability that a word is assigned to the wrong class is 0.5. Given vocabulary size V, there are on average 0.5V misclassifications. If the algorithm could inspect each word and find out its final destination, it could determine a schedule of moves that gave the minimum number of moves. The expected number of iterations needed, by an algorithm with perfect knowledge, to move the words from an initial random assignment to the correct classes, can be shown by simulation to be a little less than 0.5V. For example, when V is 1000, and the final assignment will be 700 words in one class and 300 in another, the number of moves (averaged over 100 trials) is 489. This gives a lower bound for the number of moves that the initial random assignment method will take.

The variation is that words are assigned to only one of the two possible classes at each level initially, while the other class is left empty. Then the algorithm is run as usual. The effect is that the most "alien" word (least likely to be a member of the original class) is moved to the other class, and similar words join it in further iterations. Intuitively, the "exiled" words in the second class are expected to be a minority, so the number of iterations should be less than 0.5V. This happens in practice: if the result at a particular level is (say) a 70%/30% split, McMahon's method will have taken about 0.5V iterations, while my method will have taken only about 0.3V iterations

### 4.2 Moving multiple words in parallel

At each iteration, McMahon's method selects only one word to be moved to the opposite class. My variation is: At level L (L = 0,1,...) the algorithm is splitting C classes, where $C \leq 2^L$. Instead of moving just one word per iteration, this improvement greedily tries to move C words at once, one for each class being split. It does this by tracking the "best" word in each class, instead of just the one best word over-all.

The improvement is not so dramatic as a ratio of $2^L$ to 1 might suggest. Naturally there is no improvement at level 0, where C = 1. At later levels, the maximum number (C) of words will be moved in early iterations; however the number of words moved per iteration will decrease as classes with few members have finished moving about, and will tail off to only one move per iteration when only the largest class remains to be sorted out. In very rare cases, moving multiple words may actually cause the ACMI to decrease, even though each word would cause an increase if moved by itself -- this problem is detected and corrected by retracting moves one word at a time (lowest scoring word first).

### 4.3 Change in mutual information: calculation speed

In the algorithm it is necessary to measure the change in ACMI (average class mutual information) caused by moving a word from one class to another, and this must be done $O(V^2)$ times at each level. A simple method would be to calculate the old ACMI once only, and a new ACMI after each possible move. However calculating the ACMI takes $O(C^2)$ steps, where C is the number of classes at that level. McMahon's sparse data on implementation matters do not cover this topic. My implementation has already made a drastic simplifying assumption that 10 levels are sufficient, meaning that the maximum number of classes will be $2^{10}$ i.e. 1024. This allows class bigram frequencies to be held in a simple 1024x1024 array, which can be held in real memory even on the typical modern home computer. Irrespective of whether sparse data storage techniques are used or not, a simplistic ACMI calculation method would examine all the elements in a CxC matrix. However, altering the class of a word changes only two rows and only two columns in this matrix. So I calculate only the relevant rows and columns before and after the change. This takes 8(*C*-1) steps in total. My method is faster when C is 7 or more i.e. after the first two levels. It is used in all timing results reported in this paper.

## 4.4  Retaining information

Because the time for classification is said to be O($V^3$), it is necessary to work on a subset of the vocabulary (the words of highest frequency) to obtain a basic classification. Let *F* be a frequent word and *R* be a rare word. If we use only *FF* bigrams and ignore *FR*, *RF,* and *RR* bigrams, much information might be lost.

My variation is to let the rare words still join the autoclassification, but not as individual words. Instead, they are grouped together under "pseudo-word" banners by simple morphological tags and by length. For all of the timing experiments in this paper, I chose these simple morphological groups: numeric, alphanumeric, word (alphabetic with at least one vowel), acronym (alphabetic with no vowels), and "nota" (none of the above). For example, each of `123` and `987` is treated as if it were the pseudo-word `<numeric3>`, `K9` as `<alphanumeric2>`, `ARISTOTLE` as `<word9>`, and `FLRCVRNGS` (*floor coverings!?)* as `<acronym9>`.

Note that this grouping is only to avoid information loss when assigning frequent words to classes. It is not used for the actual assignment of rarer words to classes, which is done in a later step, not covered in this paper. However examining how the pseudo-words are classified can yield useful knowledge.

## 4.5  Avoid (re)clustering closed classes

After experiments on some of my data, I found that almost all the closed-class words neatly stay in one side sub-tree of the classification tree taking off a reasonable amount of class space. Considering the grammatical nature of closed-class words, it seems that using structured tags is less necessary and less useful than with open-class words.

A proposal is run the clustering algorithm twice. The first run would be only to (say) level four, clustering as many of the most frequent words (say 1000) as could be achieved in a reasonable time. The purpose of the first run would be to help find out what the closed-class words were, and to examine their functions. A four-level classification seems to be enough to separate out all sub-groups of closed-class words from the open-class words. Then the closed-class words would be manually assigned to (non-structured) classes. The second run would attempt to cluster only the frequent words that were not on the closed-class list with pre-assigned classes. This would allow the algorithm to use the whole class space where it is needed: for open-class words. The algorithm would not try to change the pre-assigned classes, but would still use the frequencies related to the pre-assigned words, so there would be no information loss.

The time required to run a McMahon-style algorithm depends on the number of words whose classes are being changed, so pre-assigning of classes should reduce the running time. This proposal is not used in any time calculation reported in this paper.

## 4.6 Robustness evaluation using Elman's grammar

Elman (1990) experimented with a small grammar of 16 non-terminal rules and 12 terminal rules, shown below.

Elman Grammar: RHS of 16 non-terminal "sentence" productions

| | |
|---|---|
| noun-human  verb-eat  noun-food | noun-human  verb-percept  noun-inan |
| noun-human  verb-destroy  noun-fragile | noun-human  verb-intran |
| noun-human  verb-tran  noun-human | noun-human  verb-agpat  noun-inan |
| noun-human  verb-agpat | noun-anim  verb-eat  noun-food |
| noun-anim  verb-tran  noun-anim | noun-anim  verb-agpat  noun-inan |
| noun-anim  verb-agpat | noun-inan  verb-agpat |
| noun-agress  verb-destroy  noun-fragile | noun-agress  verb-eat  noun-human |
| noun-agress  verb-eat  noun-anim | noun-agress  verb-eat  noun-food |

Elman Grammar: 12 terminal "word class" productions

| Word Class | Contents |
|---|---|
| noun-human | man woman girl boy |
| noun-anim | cat mouse dog man woman girl boy dragon monster lion |
| noun-inan | book rock car cookie break bread sandwich glass plate |
| noun-agress | dragon monster lion |
| noun-fragile | glass plate |
| noun-food | cookie break bread sandwich |
| verb-intran | think sleep exist |
| verb-tran | see chase like |
| verb-agpat | move break |
| verb-percept | smell see |
| verb-destroy | break smash |
| verb-eat | eat |

To generate a sentence, one of the sentence rules is chosen at random. Then for each word-class rule in the sentence, one of the allowed words is chosen. No punctuation or other end-of-sentence marker is generated. As can be seen from the rules, apart from "break", which is in two verb-classes as well as being a food-noun (!?), and "see" (which is in two verb classes), the word classes are simple: There are 6 subclasses of verbs, as listed in the table. Nouns are divided (and further subdivided) into animate (human, aggressor, other) and inanimate (fragile, food, other).

Elman's experimental results from clustering using an artificial neural network showed poor performance.

McMahon (1994) tested his clustering algorithm on a 10,000-sentence corpus generated from Elman's grammar. He reported results better than Elman's, but still not correct: in one result, human nouns and non-aggressive animate nouns are mixed and the food-noun "sandwich" is not clustered with other food-nouns "bread" and "cookie". This experiment was repeated "several" times, each time with different but imperfect results. McMahon opines that "The reasons for the weaknesses in the fine details of the classification spring from the sub-optimal strategy of the current algorithm, combined with the initial word classification state" and that "Another limitation involved the relative sparseness of the data".

I have tested my implementation of McMahon's method (M) and my variations (ZNR: non-random initialisation, and ZNRP: ZNR plus parallel moving) in a similar manner, with corpus sizes of 10,000, 2000 and 1000 sentences, with the results shown in the table on the next page.

All methods give some incorrect results with a corpus size of 1000 sentences. My variations show stable good results in different trials, down to a corpus size of 2000 sentences. All of the weaknesses shown by McMahon's results have disappeared. It seems that sparseness of data is not the problem with McMahon's own results with a 10,000-sentence corpus. At a corpus size of 10,000 sentences, my implementation of McMahon's algorithm mostly shows very good results. However in one case the result is worse than those reported by McMahon: the major verb-noun distinction is not made.

This evidence tends to indicate that the random element in McMahon's algorithm can produce random unreliable results, and that my removal of the random element produces stable reliable results.

**Results of clustering texts generated from Elman's grammar**

| Corpus Size | Corpus Ident. | Method | Run No. | Level of Error | Comment |
|---|---|---|---|---|---|
| 10000 | 10ka | ZNR | | | |
| | | ZNRP | | | |
| | | M | 1,3,4 | | |
| | | M | 2 | medium | "smash" with nouns, "plate" & "glass" with verbs |
| 10000 | 10kb | ZNR | | | |
| | | ZNRP | | | |
| | | M | 1,2,3,4 | | |
| 2000 | 2ka | ZNR | | | |
| | | ZNRP | | | |
| | | M | 1,2 | | |
| 2000 | 2kb | ZNR | | | |
| | | ZNRP | | | |
| | | M | 1 | | |
| | | M | 2 | high | much mixing of nouns and verbs |
| 1000 | 1ka | ZNR | | low | "cat" included with humans |
| | | ZNRP | | medium | "smash" with nouns, "plate" & "glass" with verbs |
| | | M | 1 | medium | "smash" with nouns, "plate" & "glass" with verbs, "girl" with cat/dog/mouse |
| | | M | 2 | | |
| 1000 | 1kb | ZNR | | | |
| | | ZNRP | | high | much mixing of nouns and verbs |
| | | M | 1,2 | | |

### 4.7 *Evaluation on real data*

The above experiments give us some helpful insights about McMahon's algorithm and my variations. However Elman's grammar is just a tiny subset of English, with a vocabulary of only 29 words. It is necessary to run tests on real data.

I have used two widely different texts:

(1) "CB", an in-confidence extract of names of business customers from a commercial database. Each "sentence" nominally consists of a "customer name" field and a "known-as name" field, but the contents taken together consist of either or both of the name of the legal entity (company, person, partnership, trust, etc) owning the business and the trading name of the business, together with (optionally) the location and miscellaneous annotations. Here are some (constructed) examples:

> 15TH ACME P/L T/A BOB'S BETTER BEEF - SYDNEY (IN LIQ'N)
> 15TH ACME PTY LTD (BOBS BETTER BEEF)
> BOBS BETTER BEEF (15TH ACME PTY. LTD.)
> AB & CD SMITH & EF JONES --- a partnership
> A. B. BROWNING --- a person
> C. D. ENGINEERING --- a trading name

The CB text has 183,502 word-tokens and a vocabulary of 26,608.

(2) "P&P", a publicly available no-cost no-copyright text, the novel "Pride and Prejudice" by Jane Austen. After removal of chapter headings, dramatis personae, etc and tokenisation, this text has 144,484 word-tokens and a vocabulary of 6,652 word-types, including punctuation.

I have examined the effect of initial assignment of words to classes by clustering the 780 most frequent words (plus 20 "pseudo-word" groups of less frequent words, giving $V = 800$) from the P&P text, and examining the ACMI achieved at the first classification level.

The non-random-assignment method split out 136 words from the 800, giving an ACMI of 0.0174. Twenty-six runs of the random-assignment method were done. The ACMI ranged from a low of 0.0121 (30% lower than 0.0174) to a high of 0.0188 (8% higher than 0.0174). About half of the results were 0.0186 or higher.

The following results are from classifying the 500 most frequent words of the CB corpus, again examining only the first classification level: The non-random classification moves 38 words which are mostly closed- class words into the second class, giving an ACMI of 0.0268. The ACMI results from 6 applications of McMahon's method ranges from 0.0273 (about 2% better) to 0.02578 (about 4% worse). Five of the six ACMI results were better.

### 4.8 Running speed

I have compared the running speed of my implementation of the basic McMahon algorithm and my improvements. All timings were obtained using an Intel 486 66 MHz PC running under PC-DOS 6.3. The software is written in C and compiled using v2.0 of the DJGPP implementation of the GNU C compiler, with options -O2 and -m486.

**Total time (in minutes) to cluster P&P text up to given level**

| V | Method | Level | | | | | | | | | |
|---|---|---|---|---|---|---|---|---|---|---|---|
| | | 1 | 2 | 3 | 4 | 5 | 6 | 7 | 8 | 9 | 10 |
| **1013** | ZNRP | 0.6 | 1.0 | 2.9 | 3.9 | 14.1 | 22.0 | 32.9 | 43.6 | 48.3 | 59.1 |
| **800** | ZNRP | 0.4 | 0.7 | 1.7 | 2.3 | 6.5 | 10.0 | 16.2 | 26.0 | 28.5 | 33.5 |
| **643** | ZNRP | 0.3 | 0.5 | 1.2 | 1.7 | 4.3 | 5.4 | 9.3 | 12.4 | 15.1 | 17.9 |
| | ZNR | 0.3 | 0.5 | 1.9 | 3.4 | 9.8 | 14.2 | 24.7 | 37.3 | 50.7 | 66.8 |
| | M | 0.8 | 2.1 | 4.6 | 9.2 | 19.5 | 31.8 | 48.1 | 63.0 | 75.9 | 86.0 |
| **512** | ZNRP | 0.2 | 0.4 | 0.8 | 1.1 | 2.7 | 4.0 | 5.0 | 8.0 | 10.1 | 13.0 |
| | ZNR | 0.2 | 0.4 | 1.3 | 2.3 | 6.2 | 11.9 | 17.7 | 24.9 | 34.8 | 43.0 |
| | M | 0.5 | 1.4 | 3.1 | 6.0 | 10.9 | 18.0 | 26.3 | 33.9 | 40.2 | 43.3 |
| **320** | ZNRP | 0.1 | 0.2 | 0.4 | 0.6 | 1.2 | 2.0 | 2.3 | 2.6 | 2.8 | 3.5 |
| | ZNR | 0.1 | 0.2 | 0.5 | 1.1 | 3.1 | 4.3 | 6.2 | 8.5 | 10.9 | 12.7 |
| | M | 0.2 | 0.5 | 1.1 | 2.5 | 4.4 | 6.5 | 9.2 | 12.1 | 14.1 | 15.0 |
| | M | 0.2 | 0.5 | 1.2 | 2.2 | 4.2 | 6.5 | 9.3 | 11.7 | 13.1 | 13.7 |

It can be seen that although the ZNR (non-random) method is faster than the McMahon method when clustering to a small number of levels, it has lost most or all of the advantage by level 10. However the ZNRP (non-random parallel) method seems to maintain its advantage; in the above results, the ratio of the speed of the ZNRP method to the speed of the McMahon method is at least 3.0, is typically about 4.0, and can be up to about 5.5. The effect on speed of the random nature of the McMahon method is shown by the last two rows in the above table, which are the fastest and slowest of 5 runs.

A time estimate formula has been derived for clustering the P&P text to 10 levels using the ZNRP method. A linear regression gives the number of seconds to cluster to 10 levels as $\ln(\text{time}) = 2.43 \ln(V) - 8.62$. Note that the exponent of V (2.43) is much less than 3. This formula predicts 5.3 hours for V=2000, 28.4 hours for V=4000, and 6.4 days for V=8000.

## 5. Conclusion

Classification of words using structured tags has some advantages over using discrete word-classes. McMahon's method uses simple bigram statistics to produce a reasonable classification. The contribution of this paper is to explain more clearly the nature of the algorithm, to show how its robustness can be improved by removing its inherent randomness, and more importantly, to improve the speed of the algorithm significantly. The result is that this word classification method is much more practically viable. Two other proposed variations offer scope for further study and potential enhancement to classification quality and speed.

Word classification can be regarded as a fundamental step for automatic or non-automatic, mono-lingual or multi-lingual natural language processing system, especially for the languages that there is no ready computational lexicon. To have a fast, efficient automatic word classification system will be really helpful.

## Acknowledgements

I would like to thank my thesis supervisor, Dr Dominique Estival, for constructive criticism, my husband, John Machin, for assistance with English expression and program debugging, and the company which owns the customer database for kindly permitting me to use it.